# A Stock Selection Method Based on Earning Yield Forecast Using Sequence Prediction Models

Jessie Sun

**Abstract**

Long-term investors, different from short-term traders, focus on examining the underlying forces that affect the well-being of a company. They rely on fundamental analysis which attempts to measure the intrinsic value an equity. Quantitative investment researchers have identified some value factors to determine the cost of investment for a stock and compare different stocks. This paper proposes using sequence prediction models to forecast a value factor-the earning yield (EBIT/EV) of a company for stock selection. Two advanced sequence prediction models-Long Short-term Memory (LSTM) and Gated Recurrent Unit (GRU) networks are studied. These two models can overcome the inherent problems of a standard Recurrent Neural Network, i.e., vanishing and exploding gradients. This paper firstly introduces the theories of the networks. And then elaborates the workflow of stock pool creation, feature selection, data structuring, model setup and model evaluation. The LSTM and GRU models demonstrate superior performance of forecast accuracy over a traditional Feedforward Neural Network model. The GRU model slightly outperformed the LSTM model.

## 1. INTRODUCTION

Stock selection means picking individual stocks (or equities) for an investment portfolio. There are many ways to pick stocks. Traditionally, short-term traders often use technical analysis, which focuses on patterns within stock charts as a way to forecast future pricing and volume trends. In a short term, from a few minutes to several weeks, stock price can fluctuate dramatically due to the volatile nature of the stock market. Many factors, for example, company news, consumer's attitudes and etc., can all have major effects on the stock price. And by the time these types of information come out, the markets have already responded and most of the investors' gaining opportunities are gone. In contrast, the long-term investors believe a company's true value should reflect its financial performance which is reported in the company's financial statements. So for long-term investors, they have relied on fundamental analysis which focuses on evaluation of securities from available information with a particular emphasis on financial accounting information [1]. Financial accounting information is the information contained in the financial reports filed by companies. For example, in the US, all corporations that are either listed on NASDAQ or have more than 10 million asset and 500 shareholders have to file quarterly reports (10Q) to the Securities and Exchange Commission. The accounting information determines the true value of a company. If the true value is known, it can be compared with the market value for the same company as implied by the current stock price and the number of shares outstanding. And the essence of the so called value investment is to buy the most undervalued stocks and sell the most overpriced stocks. Fundamental analysis focuses on two types of numerical data presented in corporate finance report-the income statement data and the balance sheet data. These variables are referred to as fundamental accounting variables [2]. Quantitative investment researchers have identified some value factors, i.e. ratio of fundamental data to stock price, to determine the cost of investment for a stock and compare different stocks. It is based on the belief that stocks with low prices relative to their fundamentals are expected to appreciate and will properly reflect the real value of the company in the future. One value factor is known as the earning yield (EBIT/EV) which is calculated by the earnings before interest and taxes (EBIT) over the enterprise value (EV) [3]. It is considered to be a more comprehensive and accurate representation of the company's value than the market capitalization which only reflects the company's common equity. The EBIT/EV ratio provides a better comparison than the commonly used conventional profitability ratios like Return on Equity or Return on Invested Capital because of the following two reasons: 1. using EBIT instead of net income as a measure of profitability eliminates the potentially interference from tax rate differences. 2. EBIT/EV normalizes the effects of different capital structures. When comparing earnings yields, EBIT put companies with different levels of debt on an equal footing and EV takes into account the value of debt as well as the market capitalization [4].

Artificial neural network (ANN) models can be used to predict the future values according to the past training data. It focuses on the recognition of patterns and regularities embedded in data. It requires a set of training data consisting of a set of instances that have been properly labeled with the correct output. A learning procedure then generates a model that attempts to meet two objectives: match the inputs and outputs as well as possible on the training data and generalize as well as possible to new data. Feed-forward Neural Networks (FNNs) and Recurrent Neural Networks (RNNs) are the commonly used ANN models for regression-type problems. Quantitative trading researchers have been using ANN models to predict stock market performance. [5, 6] evaluated FNNs for stock market prediction. [7, 8] studied a special type of RNN – the Long Short-term Memory (LSTM) Network for stock price and return predictions. [9] used a neural tensor network to extract events from news text and then used a convolutional neural network to model the influences of these events on stock price movements. However, all these studies only evaluated one type of ANNs and were limited to a few stocks for a short time span which is not suitable for long-term investment forecast.

This paper proposes a method that predicts company's future EBIT/EV using sequence prediction models and invests in the companies with the highest rankings of EBIT/EV. In this paper, two sequence prediction models are applied. They are Long Short-term Memory (LSTM) Network and Gated Recurrent Unit (GRU) Network. They are also compared to the traditional non-sequential Feed-forward Neural Network (FNN). LSTM and GRU networks are special types of recurrent neural networks (RNNs). A RNN can be viewed as multiple FNNs where one FNN is connected to the next time step FNN through the hidden neurons. This makes RNN a powerful type of neural network to handle sequence dependence modeling such as time series prediction which forecasts future values according to previously observed values. Forecasting company's future EBIT/EV based on previously reported fundamental data is a time series prediction problem. The workflow of the proposed modeling method is shown in Fig.1.

This paper firstly introduces the theories of FNN, LSTM and GRU networks. It then elaborates the workflow of stock pool creation, feature selection, data structuring, model setup and model evaluation as shown in Fig. 1. The conclusions and future work are discussed in the end. Both LSTM and GRU models demonstrate superior performance of forecast accuracy compared to traditional FNN model. And the GRU model slightly outperformed the LSTM model.

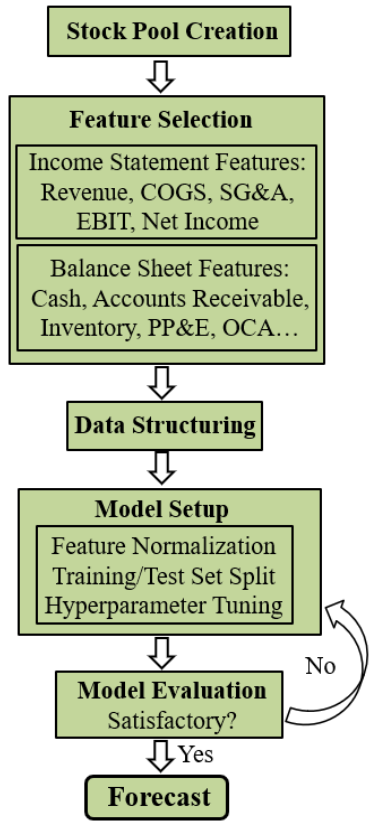

Fig.1. Workflow of the proposed modeling method

## 2. THEORIES OF THE MODELS

This section provides a brief introduction to different ANN models and serves as the mathematical foundation for the proposed method. Since LSTM and GRU models are both special types of RNN and RNNs are based on FNN, this section introduces simple FNN first and then explains the working principles of LSTM and GRU.

*2.1 Feedforward Neural Network*

An artificial neural network is a biologically inspired framework based on a collection of connected elements known as neurons. The connections between the neurons are consisted of a set of assigned weights that are adjusted in such a way that they minimize the loss of the specified objective function. FNN is the first and simplest type of ANN and a common approach to solve regression-type problems. A FNN network has three layers: input layer, hidden layer and output layer. The information moves in only forward direction (no cycles or loops) from the input neurons through the hidden neurons to the output neurons. The neurons take the values of the input data, sum them up based on the assigned weights and then add biases. By applying the activation function, the output value is determined. The operation of a neuron $P$ can be described in the following mathematical terms:

$$u_p = \sum_{i=1}^{n} w_{Pi} x_i \qquad (1)$$

$$y_p = \varphi(u_p + b_p) \qquad (2)$$

where $x_1, \ldots, x_n$ are the input variables; $w_{P1}, \ldots, w_{Pn}$ are the connection weights of neuron $P$; $u_p$ is the input combiner; $b_p$ is the bias; $\varphi$ is the activation function; and $y_p$ is the output of the neuron. [6]

*2.2 Long Short-term Memory (LSTM) Network*

LSTM network is a special type of recurrent neural networks (RNNs), i.e., neural networks of which the neurons are organized into successive layers. A RNN can be viewed as a group of FNNs where the neurons in the hidden layer of one time step are connected to every other neurons in the next successive layer [10]. This structure makes RNN suitable for dealing with time series prediction problem as it is not only based on the inputs from one time step but also takes the sequential information into account. However on the other hand, this nature of RNNs causes inherent problems known as vanishing and exploding gradients.

In order to overcome these problems, researchers proposed specially designed units for RNNs and LSTM is the first one. It was introduced by S. Hochreiter and J. Schmidhuber [11] and was further improved by researchers in the following years, for example, by F. A. Gers et al. [12], A. Graves and J. Schmidhuber [13]. LSTM networks are designed specifically to learn the long-term dependencies [14]. Like a simple FNN, a LSTM network is also consisted of an input layer, one or more hidden layers and an output layer. The main characteristics of LSTM networks is contained in the hidden layers of which the values are now determined by special units known as memory cells. Memory cells are sophisticatedly designed to control the information flow from one time step to the next. Each of the memory cells has three gates to maintain and adjust its state: forget gate ($f_t$), input gate ($i_t$) and output gate ($o_t$). $\vec{X}_t$ is the input vector at time step $t$; $\vec{H}_t$ is the hidden state vector at time step $t$ and $\vec{C}_t$ is a cell state vector stored in an external memory cell. The structure of a memory cell is illustrated in Fig.2 [15].

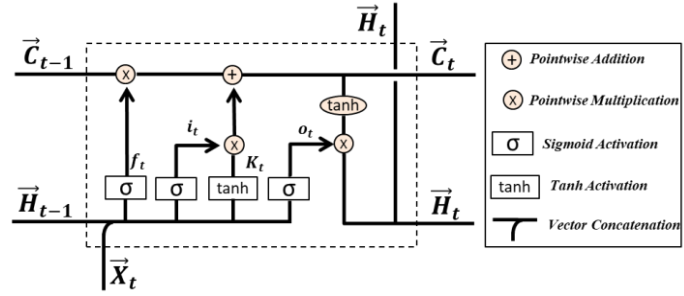

Fig.2. A LSTM unit diagram

The working theory of a memory cell is described as follows:

Firstly, the LSTM layer determines which information should be forgotten or removed from previous cell state $\vec{C}_{t-1}$. The forget gate element $f_t$ at timestep $t$ is calculated based on the current input $\vec{X}_t$, the hidden state vector $\vec{H}_{t-1}$ of the memory cell from the previous timestep $t-1$ and the bias $b_f$ of the forget gate. The sigmoid function $\sigma$ restrains all activation values into a range between 0 (completely removing the element) and 1 (completely keeping the element):

$$f_t = \sigma(\vec{W}_f \cdot [\vec{H}_{t-1}, \vec{X}_t] + b_f) \qquad (3)$$

Secondly, the LSTM layer determines what information should be added to the cell state. This step comprises operations to compute two values: a temporary cell state element $K_t$ that could potentially be added to the cell state and the input gate element $i_t$:

$$K_t = tanh(\vec{W}_k \cdot [\vec{H}_{t-1}, \vec{X}_t] + b_k) \qquad (4)$$
$$i_t = \sigma(\vec{W}_i \cdot [\vec{H}_{t-1}, \vec{X}_t] + b_i) \qquad (5)$$

Thirdly, the new cell state $\vec{C}_t$ is calculated based on the results of the previous two steps:

$$\vec{C}_t = \vec{f}_t \otimes \vec{C}_{t-1} + \vec{\iota}_t \otimes \vec{K}_t \qquad (6)$$

Finally, the hidden state vector at time step $t$ $\vec{H}_t$ of the memory cell is derived from the following two equations:

$$o_t = \sigma(\vec{W}_o \cdot [\vec{H}_{t-1}, \vec{X}_t] + b_o) \qquad (7)$$

$$\vec{H}_t = \vec{o}_t \otimes tanh(\vec{C}_t) \tag{8}$$

The input at each timestep is processed by the network as described in the equations above. Once the last element of the sequence is processed, the final output of the whole sequence is returned. Similar to FNNs, the weights and biases are adjusted as training proceeds.

*2.3 Gated Recurrent Unit (GRU) Network*

The GRU network is a newer sequence prediction model proposed in the work of J. Chung and Bengio [16]. It is also designed to solve the vanishing and exploding gradients problems of standard RNNs. The GRU unit has an architecture similar to a LSTM unit while the differences are: First, GRU eliminates the use of memory cells and the information flow is carried by the hidden states. Second, GRU merges the forget gate and the input gate into one update gate. The structure of a GRU unit is shown in Fig.3 [15]. J. Chung and Bengio describe GRU as follows:

In the first step, a reset gate element $r_t$ is computed by:

$$r_t = \sigma(\vec{W_r} \cdot [\vec{H}_{t-1}, \vec{X}_t] + b_r) \tag{9}$$

where $[\vec{H}_{t-1}, \vec{X}_t]$ is a concatenated vector of the previous hidden state vector $\vec{H}_{t-1}$ and the new input vector $\vec{X}_t$; $\vec{W_r}$ and $b_r$ are the weight and bias.

In the second step, an update gate element $u_t$ is calculated by using the same input and activation function but different weight $W_u$ and bias $b_u$:

$$u_t = \sigma(\vec{W_u} \cdot [\vec{H}_{t-1}, \vec{X}_t] + b_u) \tag{10}$$

In the third step, a temporary element $h_t$ is computed by:

$$h_t = tanh(\vec{W_h} \cdot [\vec{r}_t \otimes \vec{H}_{t-1}, \vec{X}_t] + b_h) \tag{11}$$

In the final step, the hidden state vector $\vec{H}_t$ at the current time step *t* is calculated from the previous hidden state vector $\vec{H}_{t-1}$, the update gate vector $\vec{u}_t$ and the temporary vector $\vec{h}_t$:

$$\vec{H}_t = (1 - \vec{u}_t) \otimes \vec{H}_{t-1} + \vec{u}_t \otimes \vec{h}_t \tag{12}$$

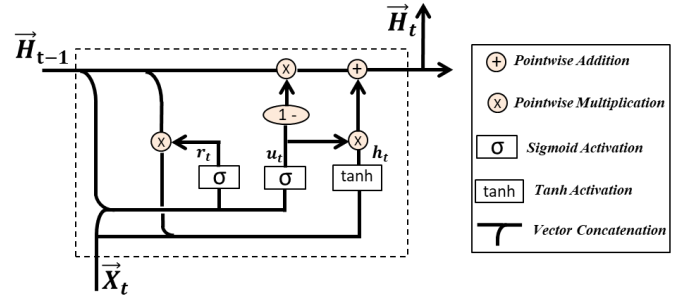

Fig.3. A GRU unit diagram

To sum up, both LSTM and GRU models have more complicated structures and more parameters than a standard RNN and FNN, which result in longer training time. However, they are both reliable sequence prediction models which solve the vanishing and exploding gradient problems of a standard RNN. Generally speaking, the GRU structure is more efficient than the LSTM structure due to fewer gates being used in the process, which requires less training time. However, from the precision perspective, GRU is not necessarily better than LSTM [16]. So for this study, both models are applied and the accuracy of them is compared.

## 3. APPLICATION

This section elaborates the workflow of stock pool creation, feature selection, data structuring, model setup and model evaluation.

*3.1 Stock Pool Creation*

In this research, SQL Server Management Studio is used to query data from S&P Capital IQ Database. 3,000 largest (ordered by the most recent Market Cap value) publicly traded US-based companies listed on NYSE, NASDAQ or AMEX are selected into the stocks pool. Financial sector companies are excluded from this research as it is a common practice in the quantitative studies to exclude financial companies because balance sheet leverage at a financial company has a very different meaning from the meaning at an operating company.

*3.2 Feature Selection*

Feature selection is normally the first step of building a neural network model [17]. By employing domain knowledge, useful raw features related to the problem are analyzed and selected. In this study, five income statement features (revenue, cost of goods sold or COGS, general and administrative expense or SG&A, earnings before interest and taxes or EBIT and net income) and nine balance sheet features (cash and cash equivalents, accounts receivable, inventory, property plant and equipment or PP&E, other current assets or OCA, debt in current liabilities, accounts payable, other current liabilities and total liabilities) are considered as input features. Detailed explanation of each feature is given below:

Revenue: the amount of money that a company actually receives during a specific period, including discounts and deductions for returned merchandise.

COGS: the direct costs attributable to the production of the goods sold in a company. It includes the cost of the materials used in creating the good along with the direct labor costs used to produce the good and excludes indirect expenses, such as distribution costs and sales force costs.

SG&A: the costs to sell and deliver products or services, in addition to the costs to manage the company.

EBIT: earnings before interest and taxes, also known as operating profit or operating income. It is an indicator of a company's profitability.

Net income: another important measure of how profitable the company is.

Cash and cash equivalents: the value of a company's assets that are cash or can be converted into cash immediately.

Accounts receivable: the amounts of money owed by customers to a company for goods or services delivered or used on credit but not yet paid for by clients.

Inventory: the goods available for sale and raw materials used to produce goods available for sale.

PP&E: long-term assets vital to business operations and not easily converted into cash.

OCA: the assets that does not include cash, securities, receivables, inventory and prepaid assets, and can be convertible into cash within one business cycle (usually a year).

Total liabilities: the money that a company owes to outside parties. It consists of current liabilities and long-term liabilities.

Other current liabilities: the current liabilities that are not assigned to common liabilities like debt or accounts payable.

Debt in current liabilities: the amount of money borrowed by one party from another.

Accounts payable: a company's obligation to pay off a short-term debt to its creditors or suppliers.

*3.3 Data Structuring*

Since many companies' fundamentals are fed into the models simultaneously and companies of different sizes usually have very different values of fundamental account variables, for normalization purpose, all raw input fundamental variables are divided by the EV of the company. As shown in Table I, the data is then constructed into a set of sequences where each sequence represents the change of a company through a period of time. The amount of high-quality data that one can currently get on broad-market company fundamentals spans about 58 years (from 1960 to 2018). It would give us 232 independent time periods (quarters) which is sufficient for learning and the time horizon is long enough not to be influenced by exogenous factors. Data is fed into the model in time steps spaced by a one-quarter interval. 14 fundamentals in two years (112 features, 8 quarters) are used to predict the company's EBIT/EV in the next quarter.

TABLE I: AN EXAMPLE OF THE DATA STRUCTURE

| Company Name | Year-quarter | Normalized Fundamental Data for the Year-quarter | | | | | | | | Next Quarter's EBIT/EV |
|---|---|---|---|---|---|---|---|---|---|---|
| | | Revenue | COGS | SG&A | EBIT | Net income | Cash and cash equivalents | … | Total Liabilities | |
| ABC Systems Corp. | 1996-Q3 | 0.58 | 0.29 | 0.19 | 0.04 | 0.03 | 0.22 | … | 0.16 | -0.03 (1998-Q3) |
| | 1996-Q4 | … | … | … | … | … | … | … | … | |
| | 1997-Q1 | … | … | … | … | … | … | … | … | |
| | 1997-Q2 | … | … | … | … | … | … | … | … | |
| | 1997-Q3 | … | … | … | … | … | … | … | … | |
| | 1997-Q4 | … | … | … | … | … | … | … | … | |
| | 1998-Q1 | … | … | … | … | … | … | … | … | |
| | 1998-Q2 | 0.94 | 0.58 | 0.30 | -0.06 | -0.04 | 0.12 | … | 0.27 | |
| | 1996-Q4 | 0.66 | 0.35 | 0.21 | 0.05 | 0.04 | 0.20 | … | 0.19 | 0.04 (1998-Q4) |
| | … | … | …. | … | … | … | … | … | … | |
| | 1998-Q3 | 1.69 | 1.02 | 0.54 | -0.03 | -0.02 | 0.21 | … | 0.51 | |
| | … | … | … | … | … | … | … | … | … | … |
| … | … | … | … | … | … | … | … | … | … | … |

*3.4 Model Setup*

This section elaborates the steps of model implementation. The normalization of features, the split of training and test set, the setup and optimization of network hyperparameters are explained as follows.

*3.4.1 Feature Normalization*

This is a necessary data preprocessing step when the input features have different units or magnitudes among them. There are many ways to normalize raw features [17]. For example, the Standard Normalization can transform the data such that its distribution has a mean value of zero and a standard deviation of one. It is given by:

$$X_{norm} = \frac{X_{Raw} - \mu}{s} \quad (13)$$

where for a specific feature, $\mu$ is the mean value of this feature and $s$ is the standard deviation of this feature.

*3.4.2 Training/Test Set Split*

The multi-time step dataset has to be randomly split into a training set and a test set. The training set is used to train the model while the test set is used independently to validate the model accuracy and assess the performance of the model. A split ratio of 80% for training and 20% for testing is used in this study.

*3.4.3 Network Hyperparameter Setup and Optimization*

A network's hyperparameter is a parameter of which the value is set before the training process while other parameters of the network are determined through training. In other words, a model learns the parameters from the data under certain hyperparameters. For a neural network, there are two types of hyperparameters. One is to determine the network structure (e.g. number of hidden layers) and the other one is to determine how the network is trained (e.g. learning rate). The traditional way to determine these hyperparameters is known as grid search, which is to search exhaustively through a manually specified finite set of reasonable values. For example, the number of hidden layers ∈ {1,2,3} and the number of neurons ∈ {40,50,60} generates 9 combinations in total. Grid search goes through all these 9 combinations and outputs the one with the best performance. A list of the optimized hyperparameters used for this study is shown in Table II.

TABLE II: OPTIMIZED HYPERPARAMETERS USED FOR THE NETWORKS

| Hyperparameters related to Network Structure | |
|---|---|
| Number of Hidden Neurons | 76 |
| Number of Hidden Layers | 2 (A LSTM/GRU layer which contains hidden states and a regular neural network layer connecting to the output layer) |
| Weight Initialization | glorot_uniform (default) |
| Activation Function | Rectified Linear Unit (ReLU) |
| Hyperparameters related to Training Algorithm | |
| Learning Rate | 0.001 (default) |
| Loss Function | Mean Absolute Error (MAE) |
| Number of Epochs | 200 |
| Batch Size | 12 |

*3.5 Model Evaluation*

The mean absolute percentage error (MAPE) is chosen as the metric to evaluate the model accuracy. MAPE is commonly used for measuring prediction error and is given as:

$$MAPE = \frac{1}{m} \sum_{i=1}^{m} \left| \frac{A_i - F_i}{A_i} \right| \times 100\% \qquad (14)$$

where $m$ is the total number of test records; $F_i$ is $i_{th}$ forecasted value; $A_i$ is the $i_{th}$ actual value. A traditional FNN model is also applied and served as a comparison group to the LSTM and GRU models.

The results of the three models' MAPEs are shown in below table:

TABLE III: PERFORMANCE COMPARISON OF DIFFERENT MODELS

| Model | MAPE (%) |
|---|---|
| LSTM | 8.26 |
| GRU | 7.05 |
| FNN | 20.66 |

Both LSTM and GRU models demonstrate superior performance of forecast accuracy compared to traditional FNN model. Although in this application the MAPE result shows the GRU model slightly outperformed the LSTM model, it is not safe to draw the conclusion that the GRU model is always better than the LSTM model from the precision perspective. In practice, it is suggested using a trial and error approach to choose between LSTM and GRU models for a specific problem with a specific dataset.

## 4. CONCLUSIONS AND FUTURE WORK

This paper proposes the use of advanced sequence prediction models to forecast companies' earning yield for stock selection. Compared to previous studies which are limited to a few stocks and short time span, this method is trained based on a data pool of 3,000 stocks over a time span of 58 years. The model learns to predict the earning yield of a company which quantifies the cost of investment for a stock and compare different stocks. Investors and portfolio managers can make informative decisions about stock selection according to the predicted earning yield values. Both LSTM and GRU have demonstrated strong ability to capture and leverage the sequential characteristics of time series data to improve prediction accuracy compared to the traditional FNN model. The GRU model slightly outperformed the LSTM model.

In future work, a categorical variable indicating the industrial sector of the company can be included as another useful feature. Better results may be obtained if markets are studied by sector, as there can be stronger correlations between stocks in the same sector.